\documentclass[runningheads]{llncs}
\usepackage[T1]{fontenc}
\usepackage{graphicx}
\usepackage{graphicx}
\usepackage{latexsym}

\usepackage{xcolor}
\newcommand{\hide}[1]{}

\begin{document}
\title{Do Language Models Understand Morality? Towards a Robust Detection of Moral Content}

\titlerunning{Do Language Models Understand Morality?}
%
\author{Luana Bulla\inst{1,2}\orcidID{0000-0003-1165-853X} \and
Aldo Gangemi\inst{2,3}\orcidID{0000-0001-5568-2684} \and
Misael Mongiovì\inst{1,2}\orcidID{0000-0003-0528-5490}}
\authorrunning{Bulla et al.}
%
\institute{University of Catania, Italy \and
ISTC - National Research Council, Rome and Catania, Italy \and
University of Bologna, Italy\\ }
%
\maketitle              
\begin{abstract}
The task of detecting moral values in text has significant implications in various fields, including natural language processing, social sciences, and ethical decision-making. Previously proposed supervised models often suffer from overfitting, leading to hyper-specialized moral classifiers that struggle to perform well on data from different domains. To address this issue, we introduce novel systems that leverage abstract concepts and common-sense knowledge acquired from Large Language Models (LLMs) and Natural Language Inference (NLI) models during previous stages of training on multiple data sources.
By doing so, we aim to develop versatile and robust methods for detecting moral values in real-world scenarios.
Our approach uses the GPT-based Davinci model as a zero-shot ready-made unsupervised multi-label classifier for moral values detection, eliminating the need for explicit training on labeled data. To assess the performance and versatility of this method, we compare it with a smaller NLI-based zero-shot model. The results show that the NLI approach achieves competitive results compared to the Davinci model.
Furthermore, we conduct an in-depth investigation of the performance of supervised systems in the context of cross-domain multi-label moral value detection. This involves training supervised models on different domains to explore their effectiveness in handling data from different sources and comparing their performance with the unsupervised methods. Our contributions encompass a thorough analysis of both supervised and unsupervised methodologies for cross-domain value detection. We introduce the Davinci model as a state-of-the-art zero-shot unsupervised moral values classifier, pushing the boundaries of moral value detection without the need for explicit training on labeled data. Additionally, we perform a comparative evaluation of our approach with the supervised models, shedding light on their respective strengths and weaknesses.
\end{abstract}
\section{Introduction}
\label{sect:intro}
The detection of moral values is a critical area of research with wide-ranging implications in fields such as Natural Language Processing (NLP), social sciences, and ethical decision-making. 
Building robust automatic systems capable of predicting the nuanced moral aspects of the text remains an ongoing challenge, especially due to the limitations of current supervised systems, which often suffer from overfitting the data distribution presented in the training dataset~\cite{bulla2023towards}. This phenomenon results in the development of hyper-specialized moral classifiers that excel at handling the specific distribution of textual documents and domains to which they have been exposed during the training phase. However, when confronted with data from sources with uncertain distribution or originating from different domains, these classifiers have a considerable drop in performance~\cite{trager2022moral,liscio2022cross}. 

To address this issue, we aim to leverage the unintentional acquisition of abstract conceptions and concepts connected to the field of social value by models during previous stages of training on multiple commonsense data~\cite{asprino2022uncovering}. This can avoid the phenomena of overfitting the training data distribution and develop versatile and robust methods applicable in real-world scenarios for reliable moral value detection. Additionally, we conduct an in-depth investigation of the performance of state-of-the-art unsupervised models by comparing them to supervised systems in a cross-domain framework. This evaluation allows us to assess the robustness and versatility of systems in effectively detecting moral values in text.\footnote{All materials and code are accessible at~\url{https://github.com/LuanaBulla/Detection-of-Morality-in-Text/tree/main}}

Our novel approach introduces the use of the GPT-3-based Davinci model~\cite{brown2020language} as a zero-shot ready-made unsupervised multi-label classifier for moral values detection. This allows us to detect moral values without the need for explicit training on labeled data.
To evaluate the performance and versatility of our approach, we compare it with a smaller, more flexible zero-shot method based on Natural Language Inference (NLI). This comparative analysis demonstrates that the NLI approach achieves competitive results compared to the Davinci model.
Furthermore, we investigate the benefits and limitations of supervised systems in cross-domain value frameworks. By training supervised models based on RoBERTa architecture~\cite{liu2019roberta} on different domains, we examine the results achieved in a cross-domain setting and compare their performance with the unsupervised systems. 
Through a detailed analysis of the results, we shed light on the strengths and weaknesses of both supervised and unsupervised methodologies, offering valuable tools for researchers and practitioners.
We consider for our study the Moral Foundation Reddit Corpus (MFRC)~\cite{trager2022moral}, which contains  $16,123$ Reddit comments split into three different sub-corpora belonging to different domains and tagged with Graham and Haidt's Moral Foundation Theory (MFT)~\cite{graham2013moral}. Based on the work of Trager et al.~\cite{trager2022moral}, we calculate the agreement between annotators to estimate the moral values associated with each comment in the sub-corpora.

The main contribution of this paper can be summarized as follows:
\begin{itemize}
\item We provide a comprehensive analysis of supervised (cross-domain) and unsupervised methodologies for moral value detection.
\item We present a novel method that leverages the GPT-3-based Davinci model as a zero-shot ready-made unsupervised moral values classifier.
\item We compare the approach based on GPT-3 with a smaller and more versatile NLI zero-shot-based method, that achieves competitive results compared with the Davinci model. 
\item We train different supervised RoBERTa models, one for each sub-corpus of the MFRC. Our findings demonstrate the remarkable effectiveness of these models in a cross-domain context, establishing a new state-of-the-art for MFRC. 
\end{itemize}

The paper is organized as follows. 
Section~\ref{sect:related} provides a summary of the current state-of-the-art results in this field. In Section~\ref{sect:theoretical}, we briefly describe the theoretical grounding of MFT. Section~\ref{sect:method} focuses on the unsupervised and supervised methodologies we employed.
In Section~\ref{sect:results}, we present an overview of our experimental settings (Sect.~\ref{sect:dataset}) and results (Sect.~\ref{sect:res}).
Section~\ref{sect:discussion} discusses the aforementioned results, comparing the unsupervised methods both internally and against the supervised systems in cross-domain settings. Finally, Section~\ref{sect:conclusion} concludes the paper and discusses potential future developments of our approach.

\section{Related Works}
\label{sect:related}
In the field of identifying moral values within text, previous research has predominantly focused on two main approaches: word-count-based methods~\cite{fulgoni2016empirical} and feature-based methods utilizing word embeddings and sequences~\cite{garten2016morality,kennedy2021moral}. These methods can be broadly categorized as supervised and unsupervised approaches. While the former relies on systems supported by external framing annotations, the latter does not require any specific external framing annotations. Unsupervised methods have explored architectures like the Frame Axis technique~\cite{kwak2021frameaxis}, demonstrated in works by Mokhberian et al.~\cite{mokhberian2020moral} and Priniski et al.~\cite{priniski2021mapping}. These approaches project words onto micro-frame dimensions, characterized by two opposing sets of words. Additionally, some unsupervised approaches leverage the extended version of the Moral Foundation Dictionary (MFD)~\cite{hopp2021extended}, which comprises words related to the virtues, vices, and general morality aspects of the five dyads of MFT. Kobbe et al.~\cite{kobbe2020exploring} contribute by linking MFD entries to WordNet, thus extending and disambiguating the lexicon within a dictionary-based framework.
Hulpus et al.~\cite{hulpus2020knowledge} propose an unsupervised approach exploring the capture of moral values through Knowledge Graphs (KG), which integrate data using a graph-structured data model. Their study evaluates the relevance of entities within WordNet 3.1, ConceptNet, and DBpedia in relation to MFT.
In recent advancements, Asprino et al.~\cite{asprino2022uncovering} introduce two distinct unsupervised approaches for detecting moral content in natural language. The first method uses KG to identify moral values within the text. By employing this approach, researchers aim to capture explicit references to moral concepts and values in the content. The second approach adopts a zero-shot machine learning model based on an NLI system as a zero-shot classifier. Authors leverage the commonsense knowledge of the NLI system, fine-tuned on a commonsense dataset for language inference (MNLI)~\cite{williams2017broad}. The indirect acquisition of commonsense knowledge by the NLI model enables the detection of main values in the input text solely based on the semantic interpretation conveyed from the MFT label taxonomy. Additionally, the authors demonstrate that incorporating the emotional tone expressed in the input sentence can significantly improve the classifier's performance. Both methods are evaluated on the Moral Foundation Twitter Corpus (MFTC)~\cite{hoover2020moral}, a dataset comprising $35$k items divided into seven domains spanning various socio-political areas, annotated with the value taxonomy of the MFT.
While these unsupervised methods offer valuable insights into the detection of moral values based on the MFT taxonomy, they do not provide a comprehensive cross-domain analysis. Furthermore, they lack a comparison with current state-of-the-art unsupervised methods represented by GPT-3.

From the cross-domain perspective, an interesting work is the Tomea system~\cite{liscio2023does}, which is an explainable method for comparing a supervised text classifier's representation of moral rhetoric across different domains. This approach aims to understand whether text classifiers learn domain-specific expressions of moral language, shedding light on differences and similarities in moral concepts across social domains. Tomea utilizes the SHapley Additive exPlanations (SHAP) method~\cite{lundberg2017unified}, which uses the Shapley values to quantify the extent to which an input component (a word) contributes toward predicting a label (a moral element), to compile domain-specific moral lexicons, facilitating direct comparisons of linguistic cues for predicting morality across diverse domains.
Additionally, other studies focus on supervised methods for cross-domain analysis, such as~\cite{liscio2022cross},~\cite{van2020recognising}, and~\cite{huang2022learning}. The first evaluates the performance of supervised systems for cross-domain value detection using the MFTC. The second focuses on the performance of a supervised model trained on non-extremist Twitter data and tested on data from extremist forums, representing a specific case of supervised cross-domain moral classification. The results demonstrate that cross-domain classification is feasible, albeit with some reduction in performance. The study compares the efficacy of Word2Vec and BERT embeddings, with BERT exhibiting slightly better generalization capabilities.
The third describes Learning to Adapt Framework (L2AF), a framework for addressing domain variations in the morality classification task when the training and testing data come from different domains. L2AF consists of four main modules: neural feature extractor (utilizing models like RNN and BERT), prediction network (predicting moral values), weighting network (adapting to domain shifts), and joint optimization. The neural feature extractor encodes input documents into feature representations. The prediction network predicts moral values using a fully connected network and softmax function. The weighting network adapts to domain shifts by assigning higher weights to out-domain instances with similar language usage to in-domain data. Joint optimization involves two tasks: moral value and domain predictions, with separate optimizers for the prediction network and weighting network. The framework dynamically adjusts and balances multi-domain data, ensuring effective adaptation to domain variations in morality classification. The authors test their framework on the MFTC, demonstrating superior performance compared to the state of the art in a cross-domain context on the same dataset. While the method shares similarities with our task and displays a high level of innovation, it primarily approaches the task from a supervised perspective and requires a substantial amount of training data, making it unfeasible when such data is not available.
In this context, the work of Guo et al.~\cite{guo2023data} addresses the recognition of moral sentiment in textual content, enabling researchers to gain insights into the role of morality in human life. Various ground truth datasets annotated with moral values have been released, each differing in data collection methods, domains, topics, and annotation instructions. Merging these heterogeneous datasets during training can result in models that struggle to generalize effectively. To overcome this limitation, the paper introduces a data fusion framework that employs adversarial domain training to align datasets into a shared feature space, thereby enhancing model generalizability. Additionally, the proposed approach uses a weighted loss function to account for differences in label distribution across datasets. As a result, the study achieves cutting-edge performance in morality inference across different datasets compared to previous methods.
While these works provide valuable insights, our focus remains on exploring effective unsupervised methods that are naturally portable across domains, unaffected by specific training data that might bias their performance. Furthermore, our study offers a comprehensive comparison of current unsupervised methods, including GPT-3 and NLI, with the performance of existing supervised approaches across domains, to verify their adaptability and robustness.

\section{Theoretical Grounding} 
\label{sect:theoretical}
Our work focuses on Haidt’s MFT~\cite{graham2013moral}, which represents the theoretical grounding for our research. The MFT is grounded
on the idea that, while morality may vary across different geographical, temporal, and cultural contexts, there are recurring patterns in its core principles, forming a psychological system of ``intuitive ethics''~\cite{graham2013moral}. MFT adopts a nativist, cultural-developmental, intuitionist, and pluralist approach to studying morality.
It acknowledges neurophysiological bases for moral responses (nativist), considers environmental influences on moral beliefs (cultural-developmental), suggests that moral judgments result from various patterns (intuitionist), and allows for multiple narratives to explain moral reasoning (pluralist). Central to MFT are six dyads of values and violations representing distinct moral dimensions.
At the core of MFT lie six dyads of values and violations, each representing distinct MFT dimensions:

\begin{itemize}
\item \textbf{Care/Harm}: This dyad is concerned with caring and harming behaviors, encompassing virtues such as gentleness, kindness, and nurturance.

\item \textbf{Fairness/Cheating}: Grounded in social cooperation and reciprocal altruism, this foundation underlies ideas of justice, rights, and autonomy.

\item \textbf{Loyalty/Betrayal}: Based on the benefits of cohesive coalitions and the rejection of traitors, this dyad emphasizes loyalty and betrayal in social interactions.

\item \textbf{Authority/Subversion}: Focusing on societal hierarchies, this foundation relates to concepts of leadership, deference to authority, and respect for tradition.

\item \textbf{Purity/Degradation}: Derived from the psychology of disgust, this dyad is associated with notions of elevated spiritual life, often expressed through metaphors like ``the body as a temple'' and incorporating spiritual aspects of religious beliefs.

\end{itemize}

These five MFT dimensions form the basis of MFT, providing a comprehensive framework for understanding the diverse aspects of human morality. In a subsequent version, the liberty/oppression dyad was introduced. This dyad represents the desire for freedom and the feeling of oppression when freedom is denied. However, we do not use it, as it is not present in the MFRC dataset.

\section{Methodology}
\label{sect:method}
Our study encompasses the implementation of both unsupervised and supervised methodologies. In the unsupervised approach, we delve into the exploration of the LLMs GPT-3 generative model, with a specific focus on the Davinci model~\cite{brown2020language}. This system plays a central role in our research as a zero-shot ready-made unsupervised multi-label moral values classifier, enabling us to detect moral values without the need for explicit training on labeled data.
Built upon the GPT-3 architecture, the Davinci model represents a cutting-edge language model renowned for its exceptional performance across a wide range of NLP tasks. Through extensive training on a vast corpus of text data, this model has acquired the ability to generate coherent and contextually relevant responses. In our methodology, we leverage the Davinci model as a zero-shot ready-made unsupervised classifier for moral value detection. By utilizing the model's intrinsic knowledge and contextual understanding, we prompt it to predict the moral values conveyed by an input text. Specifically, we employ the prompt: ``Does the sentence [Input text] convey a moral content or not? (answer with one word: moral or not moral). If yes, based on the Moral Foundation Theory, what moral values does the text reflect? (categorize text with Care/Harm, Fairness/Cheating, Loyalty/Betrayal, Authority/Subversion, Purity/Degradation).'' This prompts the model to first assess whether an item conveys moral content and, if affirmative, to label it with one or more MFT dimensions. 

\begin{figure}[h]
    \centering
    \includegraphics[width=.7\linewidth,trim={0 30px 20px 10px},clip]{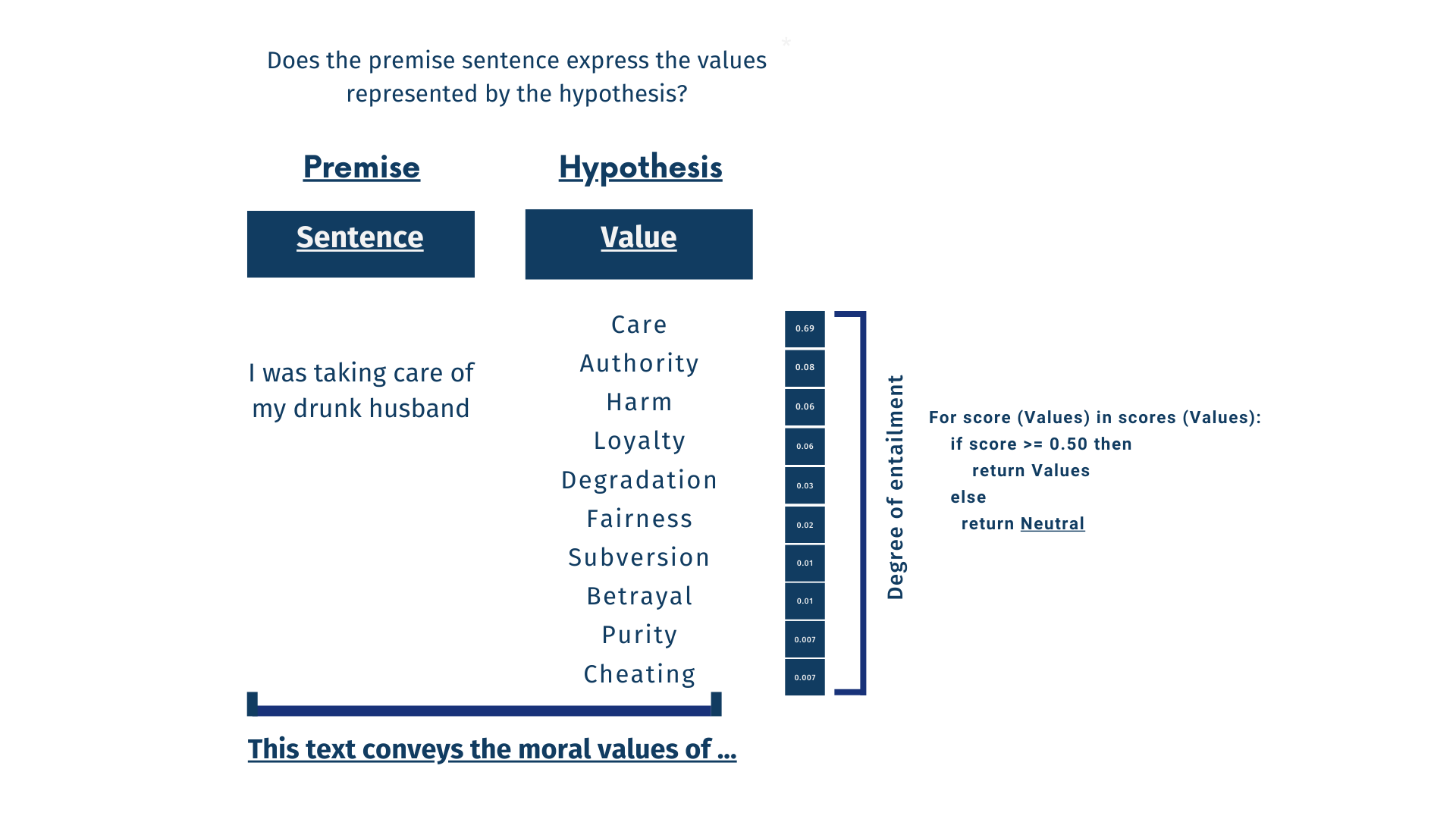}
    \caption{Overview of the NLI-based approach: The NLI system processes the input sentence as a premise and evaluates its relationship with all the labels in the MFT taxonomies, considering them as hypotheses. We select the labels with an entailment score equal to or greater than 0.50 as the classification results. If no label meets this threshold, the sentence is categorized as non-moral.}
    \label{fig:overview}
\end{figure}

As part of our exploration into unsupervised approaches for moral values detection, we delve into the capabilities of NLI systems as zero-shot classifiers. NLI models are fine-tuned to classify the relationship between two given input text: a premise and a hypothesis. Specifically, the model can accurately assess the degree of entailment, neutrality, and contradiction of the hypothesis concerning the premise. This capability allows NLI models to effectively understand the contextual relationship between textual pairs and make informed judgments about their logical connections.
Starting from the methodology proposed in a previous work~\cite{asprino2022uncovering}, we construct a hypothesis for each potential label, utilizing the input text as the NLI premise. To form these hypotheses, we employ the prompt ``This text conveys the moral values of <label>.'' Our approach considers all ten moral values and violations expressed in the MFT as possible labels.
To identify the prevalent values in the input text from an NLI perspective, we perform a multi-label classification task. Specifically, we select values with a scoring entailment output of $0.50$ or higher, indicating a strong association. Additionally, we account for neutrality, representing the items that do not convey any moral content in the dataset, by applying a $0.50$ cut-off on the normalized entailment score (for a visual representation of the methodology see Figure~\ref{fig:overview}).

In contrast to the approach defined in Asprino et al.~\cite{asprino2022uncovering}, where the entailment score for each moral value is obtained by normalizing the model's entailment and contradiction outcomes related to it, we shift our focus to the entailment and neutrality scores. We make this choice considering that the contradiction of each positive or negative label is implicitly encompassed within the range of labels we assign, taking into account the polarity of each moral foundation separately. As a result, normalizing the entailment outcome with the degree of neutrality within the same context becomes more appropriate and relevant for our analysis.
By adopting this method, we can effectively identify non-moral content through exclusion. If the degree of neutrality in a sentence exceeds the entailment for all values associated with the same item, then that item is considered to lack moral content. Consequently, the highest moral values derived from the normalization stage will fall below the 0.50 threshold. Through this approach, we enhance the precision and relevance of our moral values detection, enabling us to efficiently distinguish between moral and non-moral elements in the input text.
For our experiments, we employ the checkpoints of MNLI-RoBERTa-large\footnote{\url{https://huggingface.co/roberta-large-mnli}}, which were trained on the MNLI dataset~\cite{williams2017broad}.

Finally, we aim to explore and compare the performance of unsupervised methods based on GPT-3 and NLI with the supervised models' outcomes in a cross-domain setting. To achieve this, we implement a moral value classifier based on the RoBERTa-large model. This classifier utilize the pre-trained RoBERTa-large checkpoint\footnote{\url{https://huggingface.co/roberta-large}} and incorporate a dropout layer, a linear layer, and a softmax function applied to the pooled output embedding of the CLS token.
During training, we set the learning rate to $1e-5$, employ a batch size of $64$, and use a dropout rate of $0.1$, with AdamW as the optimizer. To evaluate the proximity of the model's predictions to the actual values, we employ the MultiLabel Soft Margin Loss\footnote{\url{https://pytorch.org/docs/stable/generated/torch.nn.MultiLabelSoftMargin} Loss.html}, a commonly used loss function for multi-label classification tasks. This evaluation allows us to assess the performance of the supervised models when faced with data from different domains, providing valuable insights into their robustness and versatility.

\section{Results and Evaluation}
\label{sect:results}
We conducted an extensive experimental analysis to assess the effectiveness of our approaches in classifying moral values, achieving a new state-of-the-art for both supervised and unsupervised methods in a cross-domain setting. In Sect.~\ref{sect:dataset} we detail the description of the MFRC while in Sect.~\ref{sect:res} we present the performance achieved by the unsupervised and supervised methods in the multi-label classification task for moral values.

\subsection{The Moral Foundation Reddit Corpus}
\label{sect:dataset}
For our experiments, we use the MFRC dataset~\cite{trager2022moral}, which enables the examination of moral language dynamics and its relationship to online and offline behaviors. Recognizing that different online platforms have distinct linguistic and social environments, the MFRC aims to address these variations by providing a diverse dataset sourced from Reddit. The corpus comprises $16,123$ Reddit comments from $12$ subreddits, which have been meticulously annotated for $8$ categories of MFT dimensions based on the updated MFT~\cite{atari2022morality}. 
In our study, we employ an earlier version of the Moral Foundation taxonomy, in which we merge the labels ``Equality'' and ``Proportionality'' into the moral category of ``Fairness''. While the Atari et al.~\cite{atari2022morality} version of the MFT separates the Fairness dimension to address distinct moral concerns related to procedural fairness and equality of outcomes, previous evidence suggests that the concept of ``Fairness'' effectively encompasses both notions. Moreover, unsupervised models demonstrate a high level of understanding and successful detection of this unified ``Fairness'' category~\cite{asprino2022uncovering}.
Furthermore, we decided not to include the label ``Thin Morality'' introduced by Trager et al., as it represents a moral judgment or concern that does not clearly align with any of the five moral dimensions, resulting in ambiguity. 
Finally, we identified items that were consistently labeled as having no moral value by the majority of annotators. Consequently, we associated these items exclusively with the ``non-moral'' label.
The corpus comprises comments from subreddits associated with US Politics, French Politics, and Everyday Moral Life. The Everyday Moral Life sub-corpus (i.e. Corpus A) encompasses topics related to various aspects of daily life, collected for their non-political moral judgment and expression of moral emotions, and includes comments from four subreddits. The US Politics sub-corpus (i.e. Corpus B) consists of comments from three subreddits that capture political moral language in general, encompassing both the moral rhetoric of the right and the left. The French Politics sub-corpus (i.e. Corpus C) contains comments from subreddits related to relevant keywords associated with the presidential race (e.g., ``Macron,'' ``Le Pen,'' ``France,'' ``French,'' and ``Hollande'). The MFRC provides information relating to the degree of confidence of the annotator for each individual item. In our study, we exclude items labeled by annotators with uncertain confidence levels and items labeled by a single annotator due to their higher subjectivity and potential noise. Consequently, the examined sub-corpora comprises a total of 
$3,472$ items labeled by two to six different annotators for the Everyday Moral Life sub-corpus; $3,949$ items labeled by two to five different annotators for the US Politics sub-corpus, and $5,443$ items labeled by two to five different annotators for the French Politics sub-corpus.
Moreover, to establish performance baselines, Reddit annotations were processed by calculating the majority vote for each moral value, considering the majority to be 50\%.

\subsection{Classification of the MFT Dimensions}
\label{sect:res}
We assess the effectiveness of both GPT-3-based and NLI-based unsupervised systems on sub-corpora A, B, and C of the MFRC dataset.
The Davinci GPT-3-based model provides a classification output that includes five distinct MFT dimensions, alongside the identification of non-moral content. Conversely, for the NLI-RoBERTa system, we present the results by converting the predictions of individual moral values or violations into their respective MFT dimensions.
To validate the performance of the supervised RoBERTa-base models in a cross-domain setting, we train them on one sub-corpus and evaluate them on the remaining sub-corpora. For this purpose, we develop three distinct supervised classifiers based on the RoBERTa-large architecture, named A-RoBERTa, B-RoBERTa, and C-RoBERTa, each trained on its respective sub-corpus: Everyday Moral Life (A), US Politics (B), and French Politics (C).
Both the unsupervised and supervised systems are tested on the complete data available in each sub-corpus. 
For an overview of the label distribution in each sub-corpora, please refer to Table~\ref{tab:support}. 

\begin{table}[]
\centering
\caption{Label distribution in each sub-corpora of the MFRC. Corpus A refers to the Everyday Moral Life sub-corpus, Corpus B indicated the US Politics sub-corpus and Corpus C represents the French sub-corpus}
\label{tab:support}
\begin{tabular}{|l|c|c|c|} \hline
\multicolumn{1}{c}{\textbf{Moral Dimension}} & \textbf{Corpus A} & \textbf{Corpus B} & \textbf{Corpus C} \\ \hline \hline
\textbf{Non-moral} & 2278 & 2684 & 4330 \\
\textbf{Fairness}  & 510  & 731  & 638  \\
\textbf{Care}      & 708  & 473  & 424  \\
\textbf{Purity}    & 102  & 90   & 75   \\
\textbf{Loyalty}   & 105  & 122  & 167  \\
\textbf{Authority} & 74   & 211  & 350  \\ \hline
\textbf{All}       & 3777 & 4311 & 5802 \\ \hline
\end{tabular}
\end{table}

Table~\ref{tab:general} presents overall results for all the methods, reporting weighted precision, recall, and F1-score for sub-corpus A, B, and C of MFRC. Among unsupervised models, the Davinci GPT-3 system achieves the highest F1 scores of 0.66, 0.59, and 0.69 for Corpus A, B, and C, respectively, while the NLI system exhibits comparable performance with F1 scores of 0.62, 0.58, and 0.69 for the corresponding sub-corpora.
Considering the supervised models, A-RoBERTa performs relatively poorer in cross-domain scenarios compared to models trained on sub-corpora B and C, with a difference of approximately ten percentage points. Notably, the supervised model trained on the USA Politics sub-corpus (B-RoBERTa) performs well in a cross-domain context, achieving F1 scores of 0.71 and 0.75 on the Everyday Life Morality (A) and French Politics (C) sub-corpora, respectively.

Tables~\ref{tab:care} to~\ref{tab:authority} report results for all MFT dimensions in terms of weighted F1-score. Davinci GPT-3 model slightly outperforms NLI system on four out of five moral dimensions, with notable differences in performance. GPT-3 exhibits a significantly better trend in the ``Fairness'' dimensions and, to some extent, in the ``Authority'' dimension, while NLI excels in ``Purity'', ``Care'', and ``Loyalty'' dimensions, consistently exceeding GPT-3 by a small margin.
Supervised models consistently outperform unsupervised models across all the MFT dimensions, aligning with overall trends observed in Table~\ref{tab:general}. Particularly, the A-RoBERTa model demonstrates better results in capturing nuances of moral values.

The behavior of supervised models is most accurately reflected in the prediction of non-moral content (Table~\ref{tab:non-moral}), which significantly influences the overall weighted F1 score reported in Table~\ref{tab:general}, where the high support of non-moral elements within the MFRC dataset plays a crucial role.
In predicting elements that do not convey any moral content, the NLI system outperforms the GPT-3-based model by 2, 6, and 5 percentage points. Supervised models demonstrate their effectiveness in classifying nonmoral elements, outperforming unsupervised models by a narrow margin of 6, 2, and 3 percentage points in corpus B, C, and A, respectively.

\begin{table}[]
\centering

\caption{Unsupervised (i.e. GPT-3 and NLI-RoBERTa) and supervised model (i.e. A-RoBERTa, B-RoBERTa, C-RoBERTa) overall performance for the task of multi-label moral values detection in terms of precision, recall and weighted F1 score. Hyphens indicate when the evaluation sub-corpus is used in training. Underlined values show the best performance by supervised models, and bold values highlight the best among unsupervised results.}
\label{tab:general}
\begin{tabular}{|l|l|c|c|c|} \hline
\textbf{Models} &
  \textbf{Metrics} &
  \textbf{Corpus A} &
  \textbf{Corpus B} &
  \textbf{Corpus C} \\ \hline \hline
GPT-3 &
  \begin{tabular}[c]{@{}l@{}}Precision\\ Recall\\ F1\end{tabular} &
  \begin{tabular}[c]{@{}l@{}}\textbf{0.64}\\ \textbf{0.72}\\ \textbf{0.66}\end{tabular} &
  \begin{tabular}[c]{@{}l@{}}\textbf{0.58}\\ \textbf{0.64}\\ \textbf{0.59}\end{tabular} &
  \begin{tabular}[c]{@{}l@{}}\textbf{0.69}\\ 0.71\\ \textbf{0.69}\end{tabular} \\ \hline
NLI-RoBERTa &
  \begin{tabular}[c]{@{}l@{}}Precision\\ Recall\\ F1\end{tabular} &
  \begin{tabular}[c]{@{}l@{}}0.60\\ 0.68\\ 0.62\end{tabular} &
  \begin{tabular}[c]{@{}l@{}}\textbf{0.58}\\ 0.63\\ 0.58\end{tabular} &
  \begin{tabular}[c]{@{}l@{}}0.68\\ \textbf{0.74}\\ \textbf{0.69}\end{tabular} \\ \hline \hline
A-RoBERTa &
  \begin{tabular}[c]{@{}l@{}}Precision\\ Recall\\ F1\end{tabular} &
  - &
  \begin{tabular}[c]{@{}l@{}}\underline{0.70}\\ 0.63\\ 0.65\end{tabular} &
  \begin{tabular}[c]{@{}l@{}}\underline{0.78}\\ 0.68\\ 0.65\end{tabular} \\ \hline
B-RoBERTa &
  \begin{tabular}[c]{@{}l@{}}Precision\\ Recall\\ F1\end{tabular} &
  \begin{tabular}[c]{@{}l@{}}0.76\\ 0.73\\ 0.71\end{tabular} &
  - &
  \begin{tabular}[c]{@{}l@{}}0.77\\ \underline{0.78}\\ \underline{0.75}\end{tabular} \\ \hline
C-RoBERTa &
  \begin{tabular}[c]{@{}l@{}}Precision\\ Recall\\ F1\end{tabular} &
  \begin{tabular}[c]{@{}l@{}}\underline{0.77}\\ \underline{0.74}\\ \underline{0.75}\end{tabular} &
  \begin{tabular}[c]{@{}l@{}}\underline{0.70}\\ \underline{0.67}\\ \underline{0.68}\end{tabular} &
  - \\ \hline
\end{tabular}
\end{table}
\vspace{1.5cm}

\begin{table}[]
\centering
\caption{Unsupervised (i.e. GPT-3 and NLI-RoBERTa) and supervised model (i.e. A-RoBERTa, B-RoBERTa, C-RoBERTa) performance for predicting the MFT \textbf{``Care''} dimension in terms of F1 score. Hyphens indicate when the evaluation sub-corpus is used in training. Underlined values show the best performance by supervised models, and bold values highlight the best among unsupervised results.}
\label{tab:care}
\begin{tabular}{|l|c|c|c|}
\hline \textbf{Models}              & \textbf{Corpus A} & \textbf{Corpus B} & \textbf{Corpus C} \\ \hline
GPT-3                & \textbf{0.54}     & \textbf{0.32}     & \textbf{0.20}      \\
NLI-RoBERTa         & 0.47     & \textbf{0.32}     & \textbf{0.20}      \\ \hline
A-RoBERTa & -        & \underline{0.48}     & \underline{0.27}     \\
B-RoBERTa & 0.53     & -        & 0.23     \\
C-RoBERTa & \underline{0.67}     & 0.46     & -       \\ \hline
\end{tabular}
\end{table}
\vspace{1.5cm}

\begin{table}[]
\centering
\caption{Unsupervised (i.e. GPT-3 and NLI-RoBERTa) and supervised model (i.e. A-RoBERTa, B-RoBERTa, C-RoBERTa) performance for predicting the MFT \textbf{``Fairness''} dimension in terms of F1 score. Hyphens indicate when the evaluation sub-corpus is used in training. Underlined values show the best performance by supervised models, and bold values highlight the best among unsupervised results.}
\label{tab:fairness}
\begin{tabular}{|l|c|c|c|}
\hline \textbf{Models}              & \textbf{Corpus A} & \textbf{Corpus B} & \textbf{Corpus C} \\ \hline
GPT-3                & \textbf{0.42}     & \textbf{0.41}     & \textbf{0.33}     \\
NLI-RoBERTa         & 0.18     & 0.17     & 0.05     \\ \hline
A-RoBERTa  & -        & \underline{0.54}     & \underline{0.50}      \\
B-RoBERTa  & \underline{0.52}     & -        & 0.45     \\
C-RoBERTa  & 0.50      & 0.52     & -       \\ \hline
\end{tabular}
\end{table}

\begin{table}[]
\centering
\caption{Unsupervised (i.e. GPT-3 and NLI-RoBERTa) and supervised model (i.e. A-RoBERTa, B-RoBERTa, C-RoBERTa) performance for predicting the MFT \textbf{``Purity''} dimension in terms of F1 score. Hyphens indicate when the evaluation sub-corpus is used in training. Underlined values show the best performance by supervised models, and bold values highlight the best among unsupervised results.}
\label{tab:purity}
\begin{tabular}{|l|c|c|c|}
\hline \textbf{Models}              & \textbf{Corpus A} & \textbf{Corpus B} & \textbf{Corpus C} \\ \hline
GPT-3                & 0.10      & 0.12     & 0.07     \\
NLI-RoBERTa         & \textbf{0.14}     & \textbf{0.14}     & \textbf{0.14}     \\ \hline
A-RoBERTa  & -        & \underline{0.20}      & \underline{0.20}      \\
B-RoBERTa  & 0.26     & -        & 0.05     \\
C-RoBERTa  & \underline{0.29}     & 0.17     & -       \\ \hline
\end{tabular}
\end{table}

\begin{table}[]
\centering
\caption{Unsupervised (i.e. GPT-3 and NLI-RoBERTa) and supervised model (i.e. A-RoBERTa, B-RoBERTa, C-RoBERTa) performance for predicting the MFT \textbf{``Loyalty''} dimension in terms of F1 score. Hyphens indicate when the evaluation sub-corpus is used in training. Underlined values show the best performance by supervised models, and bold values highlight the best among unsupervised results.}
\label{tab:loyalty}
\begin{tabular}{|l|c|c|c|}
\hline \textbf{Models}              & \textbf{Corpus A} & \textbf{Corpus B} & \textbf{Corpus C} \\ \hline
GPT-3                & \textbf{0.18}     & \textbf{0.21}     & 0.18     \\
NLI-RoBERTa         & 0.17     & 0.20      & \textbf{0.20}      \\ \hline
A-RoBERTa & -        & 0.19     & 0.24     \\
B-RoBERTa  & \underline{0.22}     & -        & \underline{0.28}     \\
C-RoBERTa  & 0.15     & \underline{0.33}     & -       \\ \hline
\end{tabular}
\end{table}

\begin{table}[]
\centering
\caption{Unsupervised (i.e. GPT-3 and NLI-RoBERTa) and supervised model (i.e. A-RoBERTa, B-RoBERTa, C-RoBERTa) performance for predicting the MFT \textbf{``Authority''} dimension in terms of F1 score. Hyphens indicate when the evaluation sub-corpus is used in training. Underlined values show the best performance by supervised models, and bold values highlight the best among unsupervised results.}
\label{tab:authority}
\begin{tabular}{|l|c|c|c|}
\hline \textbf{Models}              & \textbf{Corpus A} & \textbf{Corpus B} & \textbf{Corpus C} \\ \hline
GPT-3                & \textbf{0.16}     & \textbf{0.18}     & \textbf{0.21}     \\
NLI-RoBERTa         & 0.09     & 0.15     & 0.13     \\ \hline
A-RoBERTa  & -        & 0.26     & \underline{0.22}     \\
B-RoBERTa  & \underline{0.16}     & -        & 0.06     \\
C-RoBERTa  & 0.20      & \underline{0.29}     & -       \\ \hline
\end{tabular}
\end{table}

\begin{table}[]
\centering
\caption{Unsupervised (i.e. GPT-3 and NLI-RoBERTa) and supervised model (i.e. A-RoBERTa, B-RoBERTa, C-RoBERTa) performance for predicting the \textbf{Moral Sentiment} in terms of F1 score. Hyphens indicate when the evaluation sub-corpus is used in training. Underlined values show the best performance by supervised models, and bold values highlight the best among unsupervised results.}
\label{tab:non-moral}
\begin{tabular}{|l|c|c|c|}
\hline \textbf{Models}              & \textbf{Corpus A} & \textbf{Corpus B} & \textbf{Corpus C} \\ \hline
GPT-3                & 0.81     & 0.75     & 0.83     \\
NLI-RoBERTa         & \textbf{0.83}     & \textbf{0.81}     & \textbf{0.88}     \\ \hline
A-RoBERTa  & -        & 0.78     & 0.84     \\
B-RoBERTa  & 0.87     & -        & \underline{0.91}     \\
C-RoBERTa  & \underline{0.89}     & \underline{0.83}     & -       \\ \hline
\end{tabular}
\end{table}

\section{Discussion}
\label{sect:discussion}

The findings of our study present valuable insights into the performance of unsupervised and supervised models for classifying moral values in a cross-domain framework. Davinci GPT-3 model exhibits a slight advantage over the NLI model in the unsupervised setting, showcasing its competitiveness. This suggests that smaller and more versatile models, such as NLI, can also be effective in this domain, achieving comparable performance to LLMs.
Furthermore, GPT-3 proves to be more proficient in detecting semantic moral dimensions, particularly in opaque concepts like ``Fairness'' and ``Authority''. This advantage is attributed to GPT-3's extensive commonsense knowledge, enabling better comprehension and association with these complex moral concepts.
However, certain MFT dimensions, such as ``Authority'', ``Purity'', and ``Loyalty'', exhibit poor performance in terms of F1 score across all sub-corpora. These findings indicate the difficulty in predicting these labels, possibly due to their opaque and ambiguous meanings within the dataset's context.
On the other hand, concepts like ``Care'' and ``Fairness'' are more transparent and easily identifiable by GPT-3, and to some extent by NLI.

The training process facilitates value learning by models in a cross-domain context as models learn to discriminate moral dimensions based on the data they see during the fine-tuning phase. Nonetheless, the improvement in performance between supervised and unsupervised models is not substantial, especially for difficult-to-interpret labels. The impact of the dataset's imbalanced nature is evident, as the varying number of items assigned to each moral dimension affects the supervised models' ability to predict specific moral values, such as the moral dimensions of ``Purity'', ``Authority'', and ``Loyalty''.
Supervised models exhibit a slight advantage over unsupervised models in discriminating elements that do not convey any moral content. Indeed, while supervised models can leverage the data distribution in the training set, which includes a greater number of items labeled as non-moral, their performance does not significantly surpass that of unsupervised models in this context. This indicates that unsupervised models are more proficient at recognizing non-moral content.  

The indirect prior knowledge acquired by unsupervised models proves useful for moral detection. Particularly in predicting moral dimensions that are easier to associate at a semantic level and in discriminating items containing moral content, unsupervised models perform comparably to supervised models. This highlights the benefit of unsupervised models' general understanding of moral concepts in diverse contexts.
However, more challenging-to-interpret dyads remain an open research challenge for both supervised and unsupervised models. These ambiguous moral values present complexities in their discrimination within textual content, leading to lower accuracy in predicting them.

In conclusion, both supervised and unsupervised models exhibit strengths and limitations in classifying moral values in a cross-domain context. While GPT-3 showcases competitiveness and efficiency in detecting certain moral dimensions, challenges persist in accurately predicting difficult-to-interpret MFT dimensions. Training improves value learning, but supervised model enhancements are not consistently significant. Unsupervised models benefit from indirect prior knowledge, particularly in predicting easily traceable moral content. However, accurately discriminating opaque moral values remains a research challenge.

\section{Conclusion}
\label{sect:conclusion}
In this paper, we addressed the challenge of detecting moral values in text using both supervised and unsupervised methodologies, focusing on cross-domain scenarios.
It introduces a novel approach utilizing the GPT-3-based Davinci model as an unsupervised multi-label classifier for moral value detection and compares it with a smaller, NLI-based zero-shot method.
The study also investigates the performance of three supervised RoBERTa models trained on different sub-corpora of the MFRC dataset, establishing their effectiveness in a cross-domain context.
Through a comprehensive analysis, we compared the performance of unsupervised and supervised methodologies, shedding light on their strengths and weaknesses in cross-domain scenarios. The GPT-3-based Davinci model exhibited advantages in detecting semantic moral dimensions, especially in opaque concepts like ``Fairness'' and ``Care''. However, some moral dimensions, such as ``Authority'', ``Purity'', and ``Loyalty'', proved difficult to predict for both unsupervised and supervised models.
The dataset's imbalanced nature influenced the models' ability to predict specific moral values, particularly the challenging ones. While supervised models showed a slight advantage in discriminating non-moral elements, unsupervised models demonstrated proficiency in recognizing non-moral content.
Overall, both supervised and unsupervised models presented strengths and limitations in cross-domain moral value detection. Unsupervised models benefitted from indirect prior knowledge, while supervised models excelled in certain moral dimensions in the cross-domain setting. The challenges of accurately predicting difficult-to-interpret moral values persist for both methodologies, urging further research in this area. As part of our future work, we intend to investigate the potential use of LLMs for the generation of additional data. This approach aims to enrich the input data, to effectively address and mitigate both the labeling imbalance and subjectivity present in the dataset.

In conclusion, our study offers significant insights into the effectiveness of unsupervised and supervised models in detecting moral values across different domains. By exploring the strengths and weaknesses of both approaches, we provide researchers and practitioners with valuable tools and knowledge to enhance the development of reliable moral value detection methods. As a result, our research lays the foundation for future advancements in understanding moral content in diverse contexts, facilitating ethical decision-making, and driving progress in the fields of NLP and social sciences.

\begin{credits}
\subsubsection{\ackname} We acknowledge financial support from the H2020 projects
 TAILOR: Foundations of Trustworthy AI – Integrating Reasoning, Learning and Optimization -- EC Grant Agreement number 952215 -- and the Italian PNRR MUR project PE0000013--FAIR: Future Artificial Intelligence Research.
\end{credits}
\bibliographystyle{splncs04}
\bibliography{mybibliography}
\end{document}